# Image Retrieval Method Using Top-surf Descriptor

Ye Ji
LIACS Media Lab
jiyeanne1986@gmail.com

## ABSTRACT

This report presents the results and details of a content-based image retrieval project using the Top-surf descriptor. The experimental results are preliminary, however, it shows the capability of deducing objects from parts of the objects or from the objects that are similar. This paper uses a dataset consisting of 1200 images of which 800 images are equally divided into 8 categories, namely airplane, beach, motorbike, forest, elephants, horses, bus and building, while the other 400 images are randomly picked from the Internet. The best results achieved are from building category.

## 1. INTRODUCTION

This paper presents the results and details of a content-based image retrieval [2,3,5,7,8] research project using the Top-surf descriptor [1]. Top-surf descriptor is an image descriptor that combines interest points [7] with visual words. The project enables user search for images that contain a particular visual concept such as a car, a tree, a horse, face, etc.). Given an example image, user can drag rectangles over parts of the image which contain objects user wants and indicates these as 'positive', as well, the 'negative' can also be applied. These 'positive' and 'negative' visual words are used to search for images that contain these 'positive' visual words but 'negative' visual words. The more 'positive' visual words one image contains, the higher it will be ranked.

Top-surf descriptor is an image descriptor, which combines interest points with visual words, first presented and implemented by B. Thomee et al. in [1]. The Top-surf descriptor could result in a high performance yet compact descriptor that is designed with a wide range of content-based image retrieval applications [1]. The idea of Top-surf descriptor is based on Bag-of-Word Model. For representing one image, interest points are detected and extracted using SURF (Speeded Up Robust Features), which is a robust image detector & descriptor, first presented by Herbert Bay et al. in [2]. Then each interest point is assigned to the best match visual word according to the dictionary (codebook) and the location information is also recorded. The interest points are grouped into a number of clusters using the method based on the bag-of-words technique of Philbin et al. [3]. The dictionary contains information of the dictionary size, visual words, information of randomized Kd-Tree [4], which is used for clustering, and the idf value.

In the Top-surf, the implementation of extracting interest points is using OpenSURF library [5], which is an open-source implementation of SURF. It was quite nicely documented.

The structure of visual word contains information of index, tf, idf and location. The index actually is the index of the visual word in the dictionary. If the dictionary size is 10,000 visual words, then the indices of these visual words are from 1 to 10,000. The tf–idf weight (term frequency–inverse document frequency) is a weight often used in information retrieval and text mining, while in Top-surf, this weight is used to indicate the importance of one visual word. Location is a vector used to store the positions of the visual word in an image.

The rest of the paper is structured as follows: Section 2 describes the implementation and different features. Experiments and results are presented in section 3 and conclusions are given in section 4.

## 2. IMPLEMENTATION

The implementation of this project is started on searching images that of a certain category from the existing categories (i.e. airplane, beach, motorbike, forest, elephants, horses, bus and building), which is presented in searching images using tags. Once the user finds one image that contains something he/she is interest in, he/she drags rectangles over those interested parts. The images that fit user's search are returned.

### 2.1 Dataset

For this project, I used 8 categories: horse, elephant, beach, motorbike, bus, airplane, building and forest. In each category, there are 100 images separated into training set and testing set with 50 images respectively. In addition, 400 other images are added into the training set in order to give it some diversity. The training set was used to create the dictionary, while the testing set was used to evaluate the method. The project database contains images provided by Microsoft Research Cam-bridge and collected from the photo-sharing web-site "flickr" [6]. Use of these images must respect the corresponding terms of use.

### 2.2 Creating Dictionary

The dictionary was created from the training set (i.e. 800 images in total). I created the dictionaries by using 10,000/20,000 visual words, 25 nearest neighbors, 250 iterations and 500 random points per image. This means for each image, 500 interest points are extracted for the purpose of getting sufficient points to cluster since the size of the training set is limited, then to group all these interest points into 10,000/20,000 clusters. For each cluster, 25 nearest neighbors are determined. This process is performed 250 times so that the stability of the clusters can be ensured.

## 2.3 Preprocess of Testing Set

The images of testing set are preprocessed in order to make all the search procedure quicker. For each image, all the interest points are extracted, and the top 100 interest points are selected according to their tf-idf weight that indicates the importance. Then each interest point is assigned to which visual words it is most similar to. These visual words are saved in the file. Once it is needed, it can be loaded easily. For each visual word, all the images that contain this visual word are saved in one file under the name of this visual word's index. Furthermore, since there are 8 categories, each image is assigned a tag of its category.

## 2.4 User Interface

### 2.4.1 Search images by tag

Since there are 8 categories, this project presented a search form in which user can type in the keyword. Then all the images associate with this keyword are returned. Among these images, user can select one image that contains something he/she is looking for.

### 2.4.2 Selection

Once an image is selected, user can drag one or multiple rectangle(s) over the parts he or she interests in, and the visual words inside these rectangles are indicated as 'Positive', while the same thing can also be applied to the 'Negative' when user doesn't want the result images to contain some parts he/she doesn't like. Finally, the result images are returned in the order of how many matched visual words they contain.

## 2.5 Match Method

The Top-surf descriptor contains location information of each visual word; therefore it is quite easy to decide whether one visual word is inside in a rectangle or not. Then it is able to get the list of positive visual words as well as negative visual words. The method used in this project to match visual words is to simply compare the index of the visual word. First, get one visual word from the list; then look into the preprocessed database finding that visual word file and record the image names into a list. The more visual words one image contains, the higher it will be ranked. Since the dictionary was created by the images that belong to the same categories, they share the similar objects in these images. In addition, some diversity was added to the dictionary. Thus it is believed that the visual words in this dictionary are efficient to serve the purpose of this project.

## 3. EXPERIMENTS

In this part, 3 experiments were set to test the performance of this project. They were divided into 2 parts. In the first part, one experiment was set using the whole images, i.e. all the visual words that are extracted from the image, to get the accuracy, while the other 2 experiments were set using the selected visual words to get the accuracy. I manually selected the interest part for these experiments.

## 3.1 Using the Whole Image

Table 1 shows the accuracy of using the whole images to find images of the same category. In the first part of the experiment, the whole image was used (i.e. all the visual words) to test the performance of finding images of the same category. 50 images were used per category. I carried out this experiment by using 10,000 visual word Top-surf dictionary. The evaluation only considered the top 20 results and their rank information by using Mean Average Precision. The 20,000 visual word dictionary was also used, however, there was no big difference from using 10,000 visual word dictionary.

MAP means Mean Average Precision that is a method for evaluation of ranked retrieval results.

$$aveP = \frac{\sum_{r=1}^{N} (P(r) * rel(r))}{Number\ of\ relevant\ images}$$

where r is the rank, N the number retrieved, rel() a binary function on the relevance of a given rank, and P(r) precision at a given cut-off rank:

$$P(r) = \frac{|\{relevant\ retrieved\ images\ of\ rank\ r\ or\ less\}|}{r}$$

In Table 1, the last column shows the precision regardless of rank information. The result shows Building category performed best with 71.79% accuracy, which is far better than the result of other category. From my point of view, the reason of this is because the images of the Building category are simpler than others, which means containing fewer objects and fewer colors. The Beach category achieves the worst accuracy of less than 10%. It is because some images of beach has very few in detail, which means very few visual words could be extracted, and usually there is no specific objects.

Table 1. Accuracy of using the whole image

| Category | MAP (%) | Precision (%) |
|---|---|---|
| Airplane | 26.60% | 36.3% |
| Beach | 9.86% | 15.27% |
| Building | 71.79% | 79% |
| Bus | 28.36% | 39.7% |
| Elephant | 24.88% | 35.6% |
| Forest | 27.97% | 35.6% |
| Horse | 14.7% | 23% |
| Motorbike | 20.55% | 25.9% |

## 3.2 Using 'Positive' and/or 'Negative'

In this part, there are two parts of experiments. One is using one object in an image to detect that specific object in other images, while the other experiment is to find the images that belong to one category using part of an image of the same category.

In the first part of the experiment, I chose 2 categories to test, Airplane and Horse category. The reason why I chose these 2 categories to test is that in both of these two categories, the images are comparatively easy to decide the whether the results images fit the selected parts of retrieval images. For instance, in the images of airplane, there are two types: one is the airplane in the air while the other is the airplane on the ground. In horse category, there are brown and white horses; I could select the white horses to test the precision. The results are shown in Table 2.

There are 50 images in the airplane category, while 16 of them are the airplane in the air and the rest are the airplane on the ground. If the retrieval image is the airplane in the air, then in the result, the images that contain an airplane in the air are deemed to be correct.

For the horse category, there are 31 images that contain a white horse. During the test, I selected the white horse to retrieve images. The result images that contain a white horse are thought to be correct.

**Table 2. Accuracy of using the selection part**

| Category | MAP (%) |
|---|---|
| Airplane | 20.76% |
| Horse | 11.64% |

In the second part of this experiment, the images from 5 categories (Airplane, Horses, Elephant, Forest and Beaches) are used. The results are shown in Table 3. As for the Airplane category, I used the tail parts of the airplanes to search for the airplane images; the head of horse for the horse images; ears, trunk and ivories for the elephant images. Since there are no typical objects in the forest and beach images, I just selected random parts of images to do the experiment.

**Table 3. Accuracy of finding all the images using part of the image**

| Category | MAP (%) | Precision (%) |
|---|---|---|
| Airplane | 20.04% | 31.37% |
| Horses | 14.49% | 23.25% |
| Elephants | 18.71% | 29.25% |
| Forest | 20.79% | 30.57% |
| Beaches | 8.35% | 13% |

In this experiment, the Forest category performed the best with 20.79% accuracy, while the Beaches category performed worst with only 8.35%. The result of this experiment ranks the same as of the first experiment, but the accuracy decreases. The reason for this, I think, is less visual words were less able to describe an object in the image. Since there are specific objects in Airplane, Horses, Elephant categories, using parts of these objects can get the images in which contain similar objects; therefore, this method is likely to find the whole object in images using part of this object or the object that are similar.

## 4. CONCLUSION

In this paper, a content-based image retrieval application using Top-surf descriptor was explained and tested. It presents the searching form by query-by-example method. There are 8 categories, and the dictionary used was created also by the images of these 8 categories and some other images in order to add diversity the dictionary. Through matching visual words by the indices, the results are returned. Moreover, The user interface was designed and implemented. Although the results got from the experiments are not that satisfactory, it also shows the capability of deducing object from parts of this object or from parts of the similar objects. It can be illustrated by the experiments of using parts of the airplane, horses and elephants (the accuracies are 31.37%, 23.25%, 29.25%). For instance, as searching for the elephants images, the ears, trunk and ivories, which are unique in elephants, are used to do the searching. The accuracies got from the experiments using the whole image are higher than just using parts of the image. As for using parts of the image, it seems to perform better when the image contains less information or is less complex.